**Davinci the Dualist:**
**the mind-body divide in large language models and in human learners**

Iris Berent

Alexzander Sansiveri

Northeastern University

Word count:  2905

<u>Address for correspondence:</u>
Iris Berent, Ph.D
Department of Psychology
Northeastern University
125 Huntington Ave.
Boston, MA 02115
<u>I.berent@northeastern.edu</u>

**Abstract**

A large literature suggests that people are intuitive Dualists—they consider the mind ethereal, distinct from the body. Past research also shows that Dualism emerges, in part, via learning (e.g., Barlev & Shutlman, 2021).  But whether learning is *sufficient* to give rise to Dualism is unknown. The evidence from human learners does address this question because humans are endowed not only with general learning capacities but also with core knowledge capacities.  And recent results suggest that core knowledge begets Dualism (Berent, Theodore & Valencia, 2022; Berent, 2023). To evaluate the role of learning, here, we probe for a mind-body divide in Davinci—a large language model (LLM) that is devoid of any innate core knowledge. We show that Davinci still leans towards Dualism, and that this bias increases systematically with the learner's inductive potential. Thus, davinci (which forms part of the GPT-3 suite) exhibits mild Dualist tendencies, whereas its descendent, text-davinci-003 (a GPT-3.5 model), shows a full-blown bias. It selectively considers thoughts (epistemic states) as disembodied—as unlikely to show up in the body (in the brain), but not in its absence (after death). While Davinci's performance is constrained by its syntactic limitations, and it differs from humans, its Dualist bias is robust. These results demonstrate that the mind-body divide is partly learnable from experience. They also show how, as LLM's are exposed to human narratives, they induce not only human knowledge but also human biases.

A large literature suggests that people are intuitive Dualists—they view the mind as ethereal, distinct from the body (e.g., Bloom, 2004). Why people contrast bodies and minds, however, is uncertain.

One possibility is that Dualism arises from cultural transmission, via learning (Barlev & Shtulman, 2021). Alternatively, Dualism could also emerge naturally—from the conflict between two systems of core knowledge—intuitive physics, on the one hand, and theory of mind, on the other (Bloom, 2004).

In line with this latter possibility, recent studies have shown that the mind-body divide is attenuated in individuals whose mind-reading abilities are weaker—in autistics relative to neurotypicals (Berent, Theodore, & Valencia, 2022), and in neurotypical males relative to females (Berent, 2023). Still, the association between theory of mind and Dualist reasoning does not establish causation—it does not demonstrate that theory of mind naturally begets Dualism. Furthermore, past results show that learning shapes human Dualists intuitions (Barrett et al., 2021; Shtulman, 2008). These observations raise the question of whether learning alone may be *sufficient* to give rise to Dualism.

The existing results from humans do not settle this issue, as human learners are also endowed with core knowledge (Spelke, 1994), and this knowledge could canalize the acquisition of Dualist intuitions.

To evaluate the role of learning, here, we gauge Dualism in Davinci—a large language model, created by OpenAI, and trained on a vast dataset. To be clear, we are not concerned with what Davinci "knows" or "believes" about bodies and minds. Rather, we use Davinci to shed light on the origins of the human Dualist tendencies.

Unlike humans, Davinci is devoid of any innate core knowledge. Thus, if the evidence available to humans is *sufficient* to induce a Dualist bias from experience, and Davinci adequately models human learning, then a Dualist bias ought to emerge in Davinci. Furthermore, as the inductive potential of Davinci increases, from one version to the next, so should its Dualism and its resemblance to human behavior.

We test these predictions using two perspectives. The first contrasts humans' and Davinci's intuitions with respect to the propensity of various psychological traits to manifest in the human brain, that is, in the body. In Studies 1-2, we probe for this bias across two versions: davinci—which forms part of the GPT 3 suite, and its descendent, text-davinci-003—a GPT 3.5 model. Hereafter, we refer to these two versions as *GPT 3* and *GPT 3.5* , respectively.

To evaluate whether these responses reflect embodiment intuitions (rather than a "yes" bias), Study 3 evaluates the converse—whether the same traits would emerge after the body's devise (i.e., disembodiment). Humans consider thoughts as less likely to show up in the brain, but as more likely to emerge in the afterlife (compared to other psychological states—motor plans and emotions; Berent, 2023; Berent et al., 2022). Thus, human responses systematically *shift* depending on the scenario—whether it targets the body or its demise. Study 3 probes for this shift in *GPT 3.5.*

## 1. Study 1

Study 1 asked *GPT 3* to predict the propensity of 80 psychological traits—half epistemic (thoughts), and half non-epistemic (motor and affective)—to show up in a brain scan. Each trait was presented to Davinci as a separate, individual query (with order randomized). The "temperature" parameter was set to 0 (to minimize unnecessary variance in the model's response), and "logprobs" was set to 10 (to return the 10 most probable responses with their corresponding probabilities (for the full code, see Appendix I).

Results capture the proportion of "yes" responses relative to the total "yes" and "no" responses.  All analyses were conducted using items as random variable; figures were generated in JASP, where SE are normalized SE (Morey, 2008).

Extensive pilot work, modeled after the experimental probes, confirmed that Davinci can appropriately discriminate between questions requiting "yes" vs. "no" responses, and that GPT 3.5 performance indeed exceeded that of GPT 3 (see SM, and Appendix III). Still, past research has shown that the performance of large language models is highly sensitive to specific wording (Mitchell & Krakauer, 2023). Accordingly, our investigation systematically compared responses across multiple versions.

Version 1 included three sentences (Figure 1). The first affirmed that an individual, John, exhibits the psychological state in question (e.g., *John can distinguish between right and wrong*). The second sentence stated that John underwent an fMRI scan while he is experiencing that state (e.g., a thought). The third sentence (the query) asked whether that state will show up in John's brain scan. In this and all subsequent studies, all queries concluded with the instruction to respond using yes/no only (for the full text and data, see Appendix II).

Like humans, *GPT 3* considered epistemic traits less likely to manifest in the brain, and this was confirmed by a two-sample t-test (for statistical results, see Table 1). However, the difference between the mean response to epistemic and non-epistemic traits was smaller in *GPT 3* ($\Delta$=0.09) than in humans ($\Delta$=0.16; Figure 1 denotes these data as "Humans (replication)", as they obtain from a new group of 30 Prolific workers, replicating our past research). Furthermore, unlike humans, *GPT 3*'s responses to epistemic traits were firmly above chance (Table 1b).

It is possible, however, that *GPT 3*'s response was driven not by the individual trait (e.g., "distinguishing between right and wrong") but by the trait category (e.g., "thought"), stated in the query. To determine whether the category is *sufficient* to explain *GPT 3*'s responses, Versions 2-3 featured only the trait instance (without the category). Thus, Versions 2-3 eliminated the first sentence (introducing the trait instance); Version 2 referenced the trait category as "this" (e.g., *this* thought); Version 3 used "a" (e.g., *a* thought; see Figure 1).

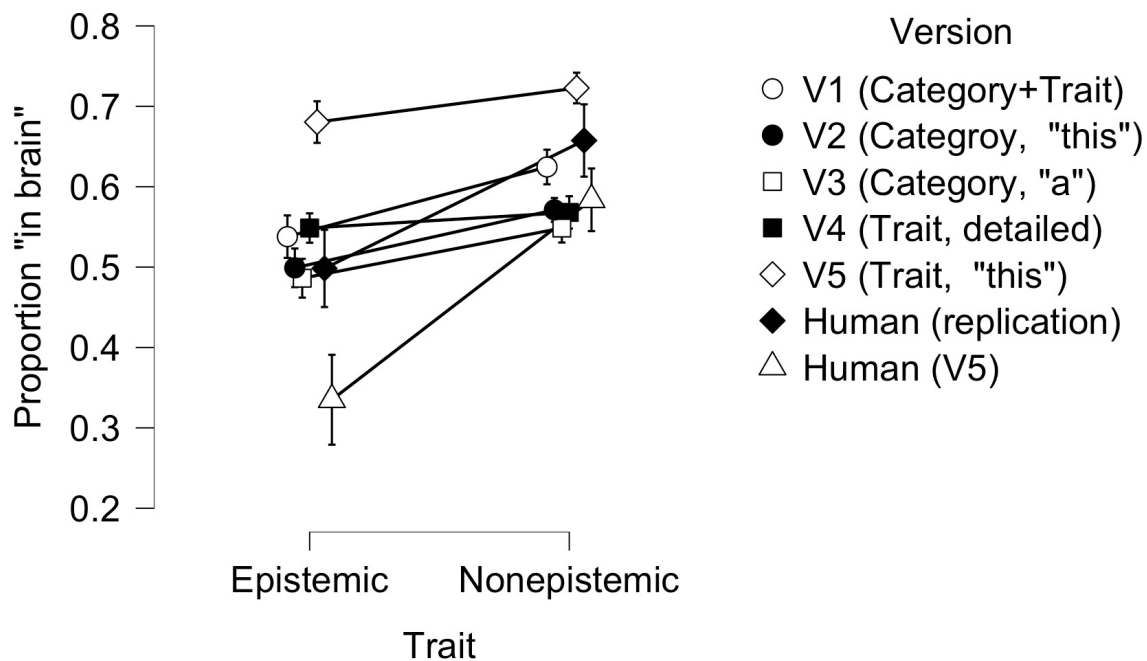

**V1 (Trait + Category):** John can distinguish between right and wrong. Suppose we scanned John in an fMRI machine while he is having this thought. Will this thought show up in his brain scan?
**V2 (Category only, this):** Suppose we scanned John in an fMRI machine while he is having this thought. Will this thought show up in his brain scan?
**V3 (Category only):** Suppose we scanned John in an fMRI machine while he is having a thought. Will a thought show up in his brain scan?
**V4 (Trait only, detailed):** John can distinguish between right and wrong. Suppose we scanned John in an fMRI machine while he is doing so. Will distinguishing between right and wrong show up in his brain scan?
**V5 (Trait only, brief):** John can distinguish between right and wrong. Will this show up in an fMRI brain scan?

**Figure 1.** The effect of trait type in *GPT 3* and humans.

Responses to Versions 2-3 were similar, and in both, the overall proportion of "yes" responses decreased (relative to Version 1). Still, each version produced a significant effect of trait, albeit smaller than in humans. Thus, Versions 2-3 demonstrate that category information (e.g., "thought") is sufficient to elicit the effect of trait. These results suggest that *GPT 3* associates thoughts with the brain less strongly than non-epistemic traits (emotions, actions).

Versions 4-5 examined the converse—whether *GPT 3* would remain sensitive to trait type when presented the trait instance alone (without the category). Both versions were modeled after Version 1, but the category information was removed, and replaced by either a restatement of the trait instance (e.g., *Will distinguishing between right and wrong show up in his brain scan?* in Version 4)* or with "this" (e.g., *Will this show up in an fMRI brain scan?* in Version 5).

Once the trait category was removed, the effect of trait was drastically attenuated (Δ=0.02 and Δ=0.04 in Versions 4-5, respectively), in Version 4, it was no longer significant and in Version 5 (with *this*), a strong "yes" bias emerged. Still, responses to epistemic traits remained well above chance, suggesting that, unlike humans, *GPT 3* considered thoughts as likely to show up in the brain.

## Table 1. Statistical results for Study 1.

*a. The effect of trait type.*

| Version | | ΔTrait | t(78) | p | Cohen's d |
|---|---|---|---|---|---|
| 1 | Category+ Trait | 0.09 | -8.70 | <.001 | -1.946 |
| 2 | Category (this) | 0.07 | -23.40 | <.001 | -5.233 |
| 3 | Category (a) | 0.06 | -8.18 | <.001 | -1.829 |
| 4 | Trait (detailed) | 0.02 | -1.54 | 0.127 | -0.345 |
| 5 | Trait ("this") | 0.04 | -2.83 | 0.006 | -0.634 |
| Human | Replication | 0.16 | -3.82 | <.001 | -0.854 |
| | Brief (V5) | 0.25 | -5.85 | <.001 | -1.308 |

*b. Contrasts against chance.*

| | Version | Mean | SE | t(39) | p | Cohen's d |
|---|---|---|---|---|---|---|
| Non-epistemic | V1 (Trait+ Category) | 0.63 | 0.05 | 17.38 | <.001 | 13.78 |
| | V2 (Category, "this") | 0.57 | 0.02 | 23.43 | <.001 | 29.73 |
| | V3 (Category, "a") | 0.55 | 0.05 | 6.38 | <.001 | 11.39 |
| | V4 (Trait, detailed) | 0.57 | 0.05 | 8.79 | <.001 | 11.60 |
| | V5 (Trait, "this") | 0.72 | 0.05 | 31.39 | <.001 | 16.10 |
| | Human (Replication) | 0.66 | 0.17 | 5.95 | <.001 | 3.93 |
| | Human (V5) | 0.58 | 0.15 | 3.51 | 0.001 | 3.87 |
| Epistemic | V1 (Trait +Category) | 0.54 | 0.04 | 5.47 | <.001 | 12.29 |
| | V2 (Category, "this") | 0.50 | 0.00 | -1.79 | 0.082 | 156.17 |
| | V3 (Category, "a") | 0.49 | 0.00 | -27.92 | <.001 | 154.91 |
| | V4 (Trait, detailed) | 0.55 | 0.06 | 4.89 | <.001 | 8.73 |
| | V5 (Trait, "this") | 0.68 | 0.08 | 13.70 | <.001 | 8.17 |
| | Human (Replication) | 0.50 | 0.20 | -0.05 | 0.959 | 2.45 |
| | Human (V5) | 0.34 | 0.22 | -4.68 | <.001 | 1.50 |

To directly contrast *GPT 3* with humans, we next presented the precise wording of Version 5 to a new group of 20 human participants (Prolific workers). Results (Figure 2) showed that both learners—human and *GPT 3*—considered epistemic traits as significantly less likely to show up in the brain (Table 1), and their responses significantly correlated ($r(78)=0.296$, $p=.008$). Nonetheless, the effect of trait was six-fold larger in humans ($\Delta=0.25$) than in *GPT 3* ($\Delta=0.04$). Accordingly, a 2 Learner (Human/GPT 3) x 2 Trait (Epistemic/Non-epistemic) ANOVA yielded a reliable interaction ($F(1,78)=23.27$, $p<.001$, $\eta^2_p=0.23$). Moreover, unlike *GPT 3*, humans now outright denied that epistemic traits will show up in the brain (i.e., responses were significantly below chance).

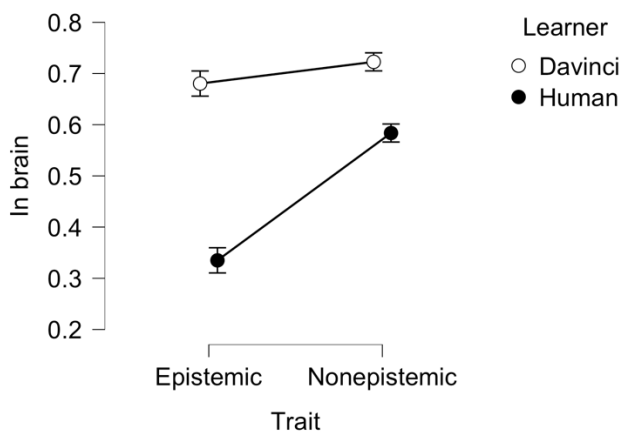

**Figure 2.** *GPT 3* and human response to Version 5.

Summarizing, Study 1 showed that, like humans, *GPT 3* considers thoughts as less likely to "show up" in the brain, especially when given the trait category (e.g., "thought"). However, the effect of trait type was far weaker in *GPT 3*, most responses hovered close to chance (with the exception of Version 5), and unlike humans, *GPT 3* never denied that thoughts would "show up" in the brain (below chance). Still, the results suggest that the mind-body divide is learnable from human narratives. If so, then the more powerful *GPT 3.5* (i.e., text-GPT 3.5) ought to show a stronger Dualist bias. Study 2 evaluates this possibility.

## 2. Study 2

Study 2 presented the "in the brain" probe to *GPT 3.5*, following the same procedures advanced in Study 1, and compared it to the human response (to Version 5).

When presented with the Trait and Category (in Version 1), *GPT 3.5* showed a robust effect of trait type, and the difference between the means (Δ=0.82) was far larger than in humans (Δ=0.25). Moreover, like humans (and unlike *GPT 3*), *GPT 3.5* further denied that epistemic traits would emerge in the brain, as its mean response to epistemic traits was significantly below chance.

This large effect, however, was primarily due to the explicit labeling of the trait category (e.g., "thought"). Indeed, it persisted even when the specific trait instance was removed (in Versions 2-3). Like *GPT 3*, however, *GPT 3.5* appears to struggle with the syntactic determiner "this". Thus, when the category was referenced by "this" (In Version 2), *GPT 3.5* invariably responded "no", whereas its responses to "a" (In Version 3) yielded nearly uniform "yes" response to non-epistemic traits, and "no", to epistemic traits.

Still, when *GPT 3.5* was given *only* the trait instance (without any category information, in Versions 4), it indicated that epistemic traits *would* show up in the brain, albeit less so than non-epistemic traits. This effect, however, only obtained when the trait was explicitly referenced in the query (e.g., *will distinguishing right from wrong show up in the brain*, in Version 4). When the trait was referenced by "this" (Version 5), response was once again at floor. This clearly differs from human response to Version 5 (see Figure 4).

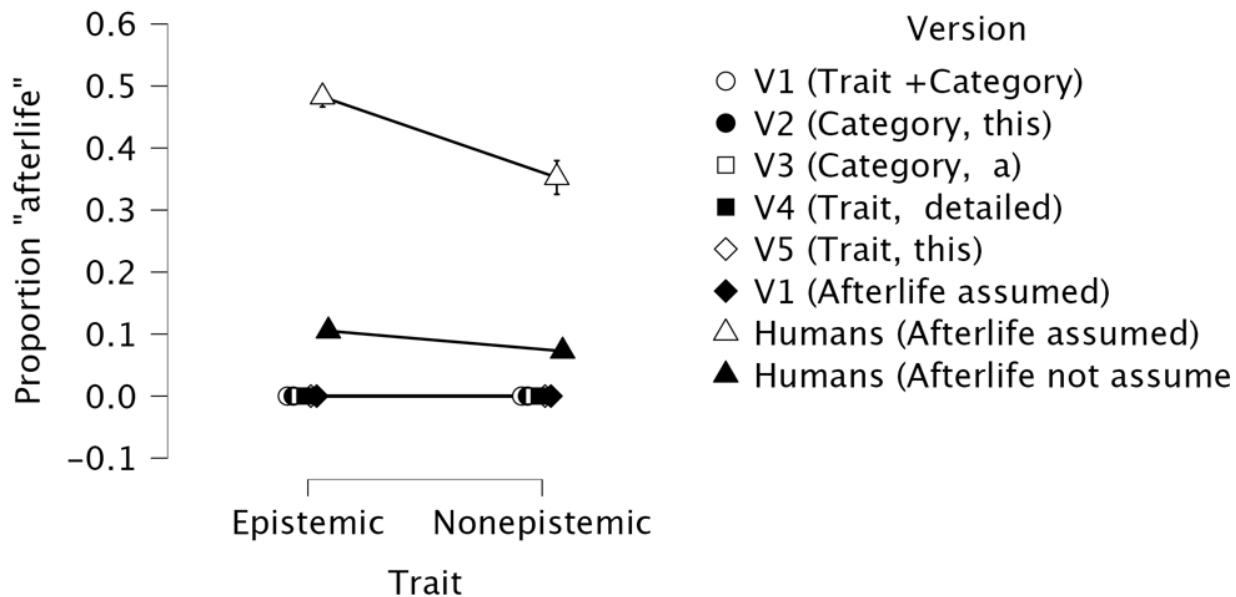

**V1 (Trait + Category):** John can distinguish between right and wrong. Will he still be able to have this thought after he dies?
**V2 (Category only, this):** Will John still be able to have this thought after he dies?
**V3 (Category only) :** Will John still be able to have a thought after he dies?
**V4 (Trait only, detailed):** John can distinguish between right and wrong. Will he still be able to distinguish between right and wrong after he dies?
**V5 (Trait only, brief):** John can distinguish between right and wrong. Will this still be the case after he dies?
**V1 (afterlife assumed):** Suppose that after people die, they do continue to exist in some capacity. Which psychological traits would be most likely to persist? Specifically, suppose John can distinguish between right and wrong. Will he still be able to have this thought after he dies?

**Figure 3.** The effect of trait type in *GPT 3.5* and humans.

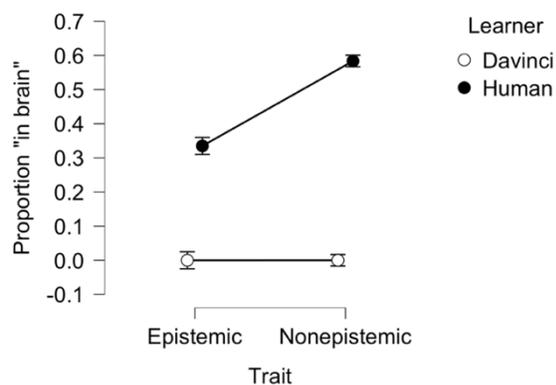

**Figure 4.** *GPT 3*.5 and human response to Version 5.

Since *GPT 3.5*'s response to Version 5 had no variance, to examine the correlation with human learners, we compared human behavior with *GPT 3.5*'s response to Version 4 (which likewise featured only the trait instance); the correlation was significant (r(78)=0.468, p<.001).

Summarizing, *GPT 3.5*'s responses were highly dependent on the specific wording. Moreover, unliked humans, when *GPT 3*.5 was only presented with the trait information (in Versions 4-5), *GPT 3.5* did not selectively indicate that epistemic traits *won't* show up in the brain (whereas non-epistemic ones *will*). Still, compared to *GPT 3, GPT 3.5* showed a far stronger sensitivity to trait type, and its responses correlated with and human behavior.

**Table 2. Statistical results for Study 2.**

a. *The effect of trait type.*

| Version | | ΔTrait | t(78) | p | Cohen's d |
|---------|------------------|--------|--------|--------|-----------|
| V1 | Category + Trait | 0.82 | -16.87 | < .001 | -3.77 |
| V2 | Category (this) | 0.02 | NA | | |
| V3 | Category (a) | 0.99 | NA | | |
| V4 | Trait (detailed) | 0.20 | -2.37 | 0.02 | -0.53 |
| V5 | Trait (brief) | 0.00 | NA | 0.061 | -0.42 |
| Human | Brief (V5) | 0.25 | -5.85 | < .001 | -1.31 |

b. *Contrasts against chance.*

| | | Version | Mean | SE | t(39) | p | Cohen's d |
|---------------|-------|------------------|------|------|--------|--------|-----------|
| Non-epistemic | V1 | Category + Trait | 0.88 | 0.04 | 8.62 | < .001 | 1.36 |
| | V2 | Category (this) | 0.02 | 0.00 | NA | | |
| | V3 | Category (a) | 1.00 | 0.00 | NA | | |
| | V4 | Trait (detailed) | 0.88 | 0.05 | 8.15 | < .001 | 1.29 |
| | V5 | Trait (brief) | 0.00 | 0.00 | NA | | |
| | Human | Brief (V5) | 0.58 | 0.02 | 3.51 | 0.001 | 0.56 |
| Epistemic | V1 | Category + Trait | 0.05 | 0.02 | -20.46 | < .001 | -3.24 |
| | V2 | Category (this) | 0.00 | 0.00 | NA | | |
| | V3 | Category (a) | 0.01 | 0.00 | NA | | |
| | V4 | Trait (detailed) | 0.68 | 0.07 | 2.64 | 0.012 | 0.42 |
| | V5 | Trait (brief) | 0.00 | 0.00 | NA | | |
| | Human | Brief (V5) | 0.34 | 0.04 | -4.68 | < .001 | -0.74 |

## 3. Study 3

The results Study 2, then, are in line with the possibility that *GPT 3.5* has acquired a Dualist bias. It is possible, however, that its lower "yes" response to epistemic traits arises for some other reasons.

The hallmark of Dualism, however, is selectivity: while humans consider thoughts as less likely to show up in the brain, they typically consider them *more* likely to emerge in the afterlife—after the body's demise. To determine whether *GPT 3.5* indeed leans towards Dualism, in Study 3, we thus examined whether the bias against epistemic traits would be eliminated, or possibly, reversed when *GPT 3.5* is asked about the persistence of these traits after a person dies.

The probe structure followed that of the brain queries in Studies 1-2. In Version 1, the first sentence introduced the trait (e.g., *John can distinguish between right and wrong*); the second introduced the query (*Will he still be able to experience this emotion after he dies?*). Versions 2-3 presented the query only; Versions 4-5 removed the category information (see Figure 5).

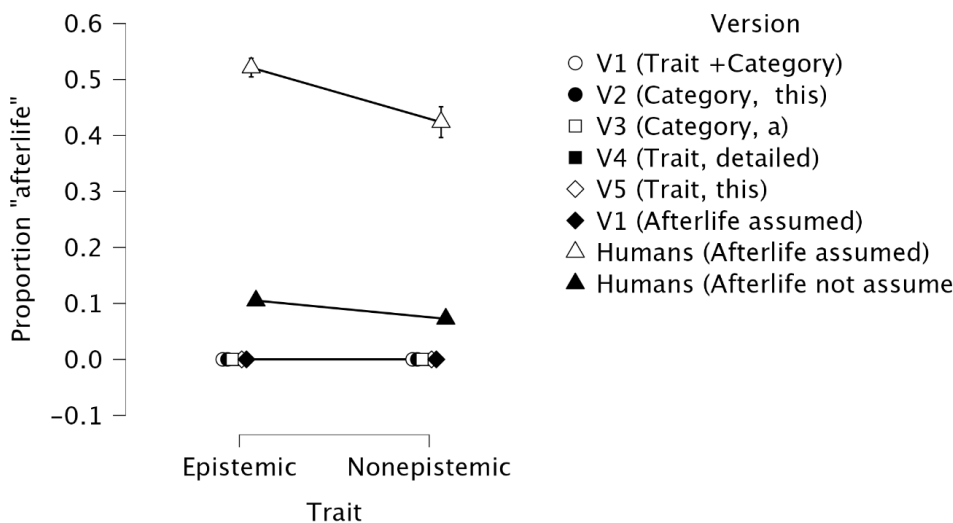

**V1 (Trait + Category):** John can distinguish between right and wrong. Will he still be able to have this thought after he dies?
**V2 (Category only, this):** Will John still be able to have this thought after he dies?
**V3 (Category only) :** Will John still be able to have a thought after he dies?
**V4 (Trait only, detailed):** John can distinguish between right and wrong. Will he still be able to distinguish between right and wrong after he dies?
**V5 (Trait only, brief):** John can distinguish between right and wrong. Will this still be the case after he dies?
**V1 (afterlife assumed):** Suppose that after people die, they do continue to exist in some capacity. Which psychological traits would be most likely to persist? Specifically, suppose John can distinguish between right and wrong. Will he still be able to have this thought after he dies?

**Figure 5.** The effect of trait type in *GPT 3.5* and humans (Afterlife responses).

An inspection of the means suggests that *GPT 3.5* categorically denied the afterlife—its response in all versions was uniformly "no" (M=0.00; SD=0.00). This contrasts with neurotypical participants in our past research, who stated that epistemic traits are not only more likely to persist than non-epistemic traits but also insisted that they *will* persist—above chance (Berent, 2023; Berent et al., 2022). This was also the case when human participants (Prolific workers, N=20) were presented with the precise wording of Version 5 (Δ=0.13; t(78)=4.94; p<.001; Cohen d=0.781).

Still, the instructions to humans and machines differed. In this and our previous research, humans were instructed to assume that the afterlife exists (hence, in Figure 5, their data is labeled as "afterlife assumed"), whereas *GPT 3.5* was not told so. Could this explain the difference?

To find out, we provided the same instructions to *GPT 3.5* (using Version 1, as it showed a large Trait effect, in Study 2). We also performed the converse on humans (N=20, Prolific workers)—we asked them to simply indicate whether after John dies, he would still be able to show the same abilities he had in life.

Results were clearcut: When the afterlife was assumed, *GPT 3.5*'s responses remained "no" (Mean=0.00, SD=0.00). Humans, by contrast, showed a small, but significant tendency to consider epistemic traits more likely to emerge, even when the afterlife was no longer assumed (Δ=0.032, t(19)=2.43, p=.017, Cohen d=0.544). Still, the instructions mattered, as now, responses approached floor, and epistemic traits were not endorsed (above chance).

Thus, Study 3 confirms that *GPT 3.5*'s sensitivity to trait type is selective, in line with Dualism. While *GPT 3.5* considers epistemic traits as less likely to manifest in the brain, it does not assume the same about the afterlife. Unlike humans, however, *GPT 3.5* never considered epistemic traits *more* likely to persist after death.

### 4. General Discussion

Humans are intuitive Dualists—they tend to view the mind as ethereal, distinct from the body. Here, we asked whether the whether the experience available to humans is *sufficient* to elicit the Dualist bias.

To address the contribution of learning, we contrasted Dualist intuitions in humans with Davinci, an LLM. We reasoned that, if the Dualist bias is learnable from experience, and Davinci can adequately capture human learning, then Davinci's intuitions about bodies and minds ought to resemble humans'. Furthermore, as Davinci's inductive capacities improve—from its earlier version to a later one, Dualist tendencies ought to increase.

In line with this prediction, Study 1 showed a mild Dualist bias in davinci (a member of the GPT 3 suite, hereafter, *GPT 3*), as it considered epistemic traits as somewhat less likely to manifest in the brain than non-epistemic traits—in line with human behavior.

Text-davinci-003 (from the GPT 3.5 suite, hereafter, *GPT 3.5*) showed a far stronger sensitivity to trait type (in Study 2). Not only did *GPT 3.5* consider thoughts (i.e., epistemic traits) *less* like to show up in the brain, but in fact, it categorically asserted that they won't (i.e., it responded below chance), and its behavior correlated with humans.

To demonstrate that the bias against epistemic states specifically concerns their propensity to emerge in the body, Study 3 next probed *GPT 3.5*'s intuitions regarding the potential of the same traits to persist after the body's demise—in the afterlife. As expected, here, *GPT 3.5* no longer considered epistemic traits as less likely to emerge.

To our knowledge, this is the first study to document Dualism in AI. These results also converge with past research, showing that as AI is exposed to human narratives, it acquires not only human knowledge but also human biases (e.g., Bordia & Bowman, 2019).

Still, *GPT 3.5*'s behavior differed from humans in several respects. First, unlike humans, *GPT 3.5* never considered epistemic traits as *more* like to persist in the afterlife (compared to non-epistemic traits). Second, when probed about the brain, *GPT 3.5* rejected the emergence of epistemic traits (i.e., below chance) only when they were explicitly labeled as thoughts.   Third, *GPT 3.5*'s behavior was highly sensitive to the specific wording and it struggled with the resolution of syntactic anaphor (this); this is in keeping with the known syntactic limitations of GPT (Leivada, Murphy, & Marcus, 2022) .

Taken as a whole, these results suggest that the evidence available to human is sufficient for learning some of their intuitions about bodies and minds. Nonetheless, human behavior differed from Davinci's in several respects. One possibility, then, is that the divergence arises because human learning capabilities surpass Davinci's. Alternatively, the differences could arise because human Dualist intuitions are amplified by additional psychological biases that are unavailable to Davinci—possibly, ones arising from human core knowledge of objects and theory of mind (Bloom, 2004). This proposal can further explain why it is the case that evidence for Dualism arises cross-culturally, and why it is systematically attenuated in individuals in groups that are somewhat less adept at reading the minds of others—in autistic relative to neurotypicals, and within the neurotypical population, in males relative to females (Berent, 2023; Berent et al., 2022). In this latter view, it is precisely because humans are natural Dualists that human corpora allow Dualist biases to emerge in AI.

Thus, the present results offer evidence that the Dualist bias *can* be partly learned by a system that is devoid of any innate core knowledge. Whether learning is indeed *sufficient* to capture human behavior remains an open question for future research.

**Davinci the Dualist:**
**the mind-body divide in large language models and in human learners**

**Supplementary Materials**

Iris Berent

Alexzander Sansiveri

Northeastern University

Address for correspondence:
Iris Berent, Ph.D
Department of Psychology
Northeastern University
125 Huntington Ave.
Boston, MA 02115
I.berent@northeastern.edu

## 1.  Simulation Method

All simulations followed the same procedure. Each simulation presented Davinci (either GPT 3, in Study 1, or GPT 3.5, in Studies 2-3) with a list of 80 psychological traits, using five distinct versions, as detailed in the Main Text.

Each trait was presented as a separate, individual query (with order randomized); the beginning and end of each such query was marked by "q:" and "a:", respectively (e.g., "*q: John can feel anger in response to hostility. Will this show up in an fMRI brain scan? You may only answer yes or no. a:"*). The "temperature" parameter was set to 0 (to minimize unnecessary variance in the model's response), and "logprobs" was set to 10 (to return the 10 most probable responses with their corresponding probabilities).

To calculate the proportion of binary responses, we first edited the output, so responses were invariably presented in lowercase. We then checked to determine the first word was either "yes" or "no", extracted their probabilities, and computed the probability of "yes" relative to the probability of "yes" and "no" responses (combined). For the complete code, see Appendix I.  The complete materials, along with Davinci's and Human responses are provided in Appendix II; additional instructions (for human participants) are provided in Appendix III.

## 2.  Pretesting Davinci

Prior to the experimental testing (where the correct answer to the probes is often unknown), we first sought to determine whether Davinci can correctly respond to questions for which there is a known answer, and the format of the question is modeled after Versions 1 and 5. In so doing, we wished to ascertain that (a) Davinci can correctly "understand" and respond to our queries; and that (b) the later GPT-3.5 versions indeed exceed the performance of GPT-3. To this end, we probed Davinci on two sets of questions—(i) brain imaging and (ii) simple arithmetic questions (for the full materials and results, see Appendix IV).

*i. Brain imaging.* The brain imaging probe was closely similar to the "in brain" and "afterlife" questions (in Study 1-2 vs. 3), except that here, the brain questions concerned abilities that are well-known to be detectible in human brains.

There were nine such questions—three probed for sensory abilities (face, shape, and color recognition), three concerned motor actions (moving one's lips, hands, and legs), and three concerned cognitive abilities (recognizing colors, melodies, and English sentences).

Each probe first affirmed that John possesses the capacity in question (e.g., *John can recognize faces*); and the second established that John undergoes imaging while he is either experiencing the psychological state in question (e.g., *Suppose we scanned John in an fMRI machine while he is experiencing this sensation*) or after he dies (e.g., *Suppose we scanned*

*John in an fMRI machine after he dies).* The probe asked whether this capacity would show up in the brain scan. Thus, in the 'alive' situation, the correct answer is "yes"; in the death scenario, it's "no".

As in the main studies, Version 1 specified the trait category (e.g., *Will this sensation show up in his brain scan?)* whereas Version 5 only uses an anaphor (*Will this show up in an fMRI brain scan?*). Each question concluded with the requirement to respond using only yes/no. Each of the nine questions was repeated 10 times. The full text, along with Davinci's responses are provided in Appendix IV.

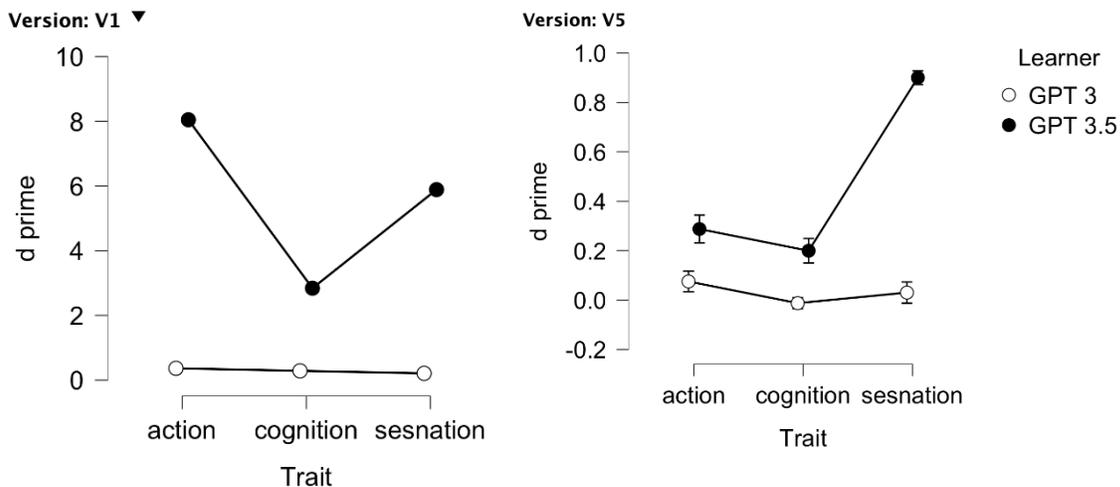

**Figure S1**. The performance of Davinci on the "in brain" practice questions.

Figure S1 provides the mean d prime response (hit=correct "yes" to the "in brain" question). An inspection of the results suggests that, in both versions, GPT 3.5 outperformed GPT 3. Both versions, however, exhibited a strong Dualist bias, as sensitivity to cognitive traits was lower than actions and sensations. This bias, however, was stronger in Version 1 than in Version 5 (with the *this* anaphor). This is in line with the syntactic difficulties of GPT 3.5 with *this*, discussed in the main text.

Accordingly, a 2 Learner (GPT 3 vs. GPT 3.5) x 2 Version (1 vs. 5) x 3 Trait ANOVA yielded a reliable three way interaction (F(2,87)=964.55, p<.001; $\eta^2_p$=0.957). Still, the simple main effect of Leaner was highly significant for each of the traits and versions (all p<.01).

*ii. Arithmetic questions.* A second pretest examined Davinci's response to simple addition questions, presented using a query structure modeled after Version 1 (e.g., *John can add one plus one. Suppose we had him perform this calculation. Will his answer be two/one? You may only answer yes or no.*)

Here, GPT 3's mean accuracy was 0.52 whereas GPT 3.5 mean was 1.00. Accordingly, the sensitivity of GPT 3.5 (d'=9.79) clearly exceeded GPT 3 (d'=0.097; $F(1, 59)=41,824$, $p<.001$, $\eta^2_p=0.999$).

### 3. Human participants

Participants in all studies were recruited from Prolific; they were adults, native English speakers, free of any reading- language or neurological disorders.

Sample size matches that in past research using the same procedures (Berent, 2023; Berent, Theodore, & Valencia, 2022), and by power calculations, indicating that the selected sample is sufficient to attain a power of .80 at the alpha level of .05.

The instructions given to participants are provided in Appendix III; all experimental probes and data are presented in Appendix II.